\documentclass[10pt,twocolumn,letterpaper]{article}

\usepackage{cvpr}
\usepackage{times}
\usepackage{amsmath}
\usepackage{graphicx}
\usepackage[utf8]{inputenc} 
\usepackage[T1]{fontenc}    
\usepackage[pagebackref=true,breaklinks=true,letterpaper=true,colorlinks,bookmarks=false]{hyperref}       
\usepackage{url}            
\usepackage{multirow}
\usepackage{amsfonts}       
\usepackage{microtype}      
\usepackage{color}
\usepackage{algorithm}
\usepackage{algorithmic}
\usepackage{rotating}
\usepackage{flushend}
\usepackage[table]{xcolor}
\usepackage{fancyhdr}
 
\cvprfinalcopy

\def\cvprPaperID{4685}

\newcommand{\N}{{\cal N}}

\title{Beyond Gradient Descent for Regularized Segmentation Losses}

\author{
   Dmitrii Marin\thanks{University of Waterloo, Canada} \thanks{Vector Research Institute, Canada} \hspace{1cm}
   Meng Tang\footnotemark[1] \hspace{1cm}
   Ismail Ben Ayed\thanks{École de technologie supérieure ÉTS, Canada}  \hspace{1cm}
   Yuri Boykov\footnotemark[1] \footnotemark[2] 
}

\begin{document}

\maketitle
\thispagestyle{fancy}

\begin{abstract}
The simplicity of gradient descent (GD) made it the default method for training ever-deeper and complex neural networks. Both loss functions and architectures are often explicitly tuned to be amenable to this basic local optimization.
In the context of weakly-supervised CNN segmentation, we demonstrate a well-motivated loss function 
where an alternative optimizer (ADM)\footnote{\url{https://github.com/dmitrii-marin/adm-seg}} achieves the state-of-the-art while GD performs poorly. 
Interestingly, GD obtains its best result for a ``smoother'' tuning of the loss function. 
The results are consistent across different network architectures. 
Our loss is motivated by well-understood MRF/CRF regularization models in ``shallow'' segmentation 
and their known global solvers. Our work suggests that network design/training should pay more attention to optimization methods.
\end{abstract}

\section{Motivation and Background} 
\label{sec:intro}

Weakly supervised training of neural networks 
is often based on {\em regularized losses} combining an empirical loss with some regularization term, which
compensates for lack of supervision \cite{weston2012deep,goodfellow2016deep}. Regularized losses are also useful for
CNN segmentation \cite{NCloss:CVPR18,Rloss:ECCV18} where full supervision is often infeasible, 
particularly in biomedical applications. 
Such losses are motivated by regularization energies in {\em shallow}\footnote{In this paper, 
``shallow'' refers to methods unrelated to deep learning.} segmentation, where multi-decade research 
went into designing robust regularization models based on geometry \cite{Mumford-Shah-89,Sapiro:97,BK:ICCV03}, 
physics \cite{Kass:88,BlakeZis:87}, or robust statistics \cite{Geman:84}. Such models should represent realistic shape priors 
compensating for image ambiguities, yet be amenable to efficient solvers. Many robust 
regularizers commonly used in vision \cite{SzeliskiMRFCompare:08,kappes2015comparative} are non-convex and 
require powerful optimizers to avoid many weak local minima. Basic local optimizers typically fail to produce practically useful results with such models.

Effective weakly-supervised CNN methods for vision should incorporate priors compensating for image data ambiguities 
and lack of supervision just as in shallow vision methods. For example, recent work \cite{weston2012deep,Rloss:ECCV18} formulated the problems 
of semi-supervised classification and weakly-supervised segmentation as minimization of regularized losses. 
This principled approach outperforms common `'proposal generation'' methods \cite{scribblesup,kolesnikov2016seed} computing ``fake'' ground truths to mimic standard fully-supervised training.  
However, we show that the use of regularization models as losses in deep learning is limited by GD, 
the backbone optimizer in current training methods. It is well-known that GD leads to poor local minima for many regularizers in shallow segmentation
and many stronger algorithms were proposed~\cite{BJ:01,BVZ:PAMI01,trws,SzeliskiMRFCompare:08,lsa-tr}. Similarly, we show better optimization beyond GD for regularized losses in deep segmentation.

One popular general approach applicable to regularized losses is ADMM \cite{Boyd2011} that splits optimization into two efficiently solvable sub-problems
separately focusing on the empirical loss and regularizer. We advocate similar {\em splitting} to improve optimization of regularized losses 
in CNN training. In contrast, ADMM-like splitting of network parameters in different layers was used  in \cite{taylor2016training} to improve parallelism.

In our work weakly-supervised CNN segmentation is a context for discussing regularized loss optimization. As a regularizer, 
we use the common Potts model \cite{BVZ:PAMI01} and consider its nearest- and large-neighborhood variants, a.k.a. sparse {\em grid CRF} and {\em dense CRF} models. 
We show effectiveness of ADMM-like splitting for grid CRF losses due to availability of powerful sub-problem solvers, \eg graph cuts \cite{BK:ICCV03}.
As detailed in \cite[Sec.3]{Rloss:ECCV18}, an earlier iterative proposal-generation technique by \cite{kolesnikov2016seed} can be related to regularized loss splitting, 
but their method is limited to dense CRF and its approximate {\em mean-field} solver~\cite{koltun:NIPS11}. In fact, given such weak sub-problem solvers, 
splitting is inferior to basic GD over the regularized loss \cite{Rloss:ECCV18}. More insights on grid and dense CRF are below.

\subsection{Pairwise CRF for Shallow Segmentation} 
\label{sec:shallow}

\begin{figure*}[t]
\centering
\begin{tabular}{ccc}
\includegraphics[width=1.7in]{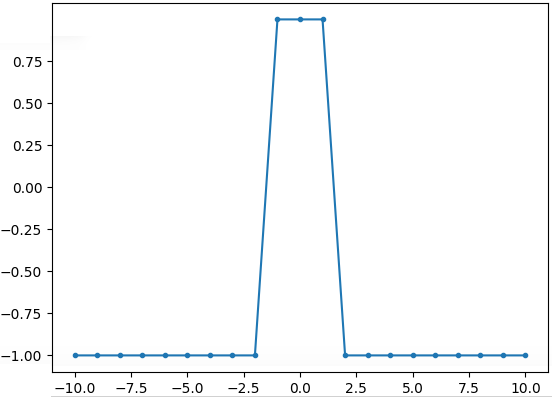}   &
\includegraphics[width=1.7in]{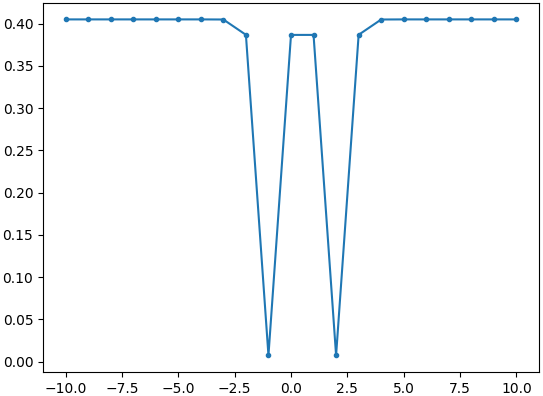}  &
\includegraphics[width=1.7in]{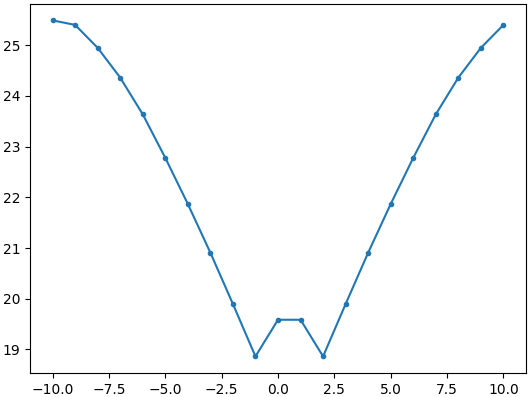}   \\
(a) 1D image & (b) grid CRF \cite{BJ:01}  & (c)  dense CRF  \cite{koltun:NIPS11}
\end{tabular}
\caption{{\em Synthetic segmentation example} for grid and dense CRF (Potts) models: (a) intensities $I(x)$ on 1D image. 
The cost of segments $S^t = \{x\;|\;x<t\}$ with different discontinuity points $t$ according to (b) nearest-neighbor (grid) Potts 
and (c) larger-neighborhood (dense) Potts. The latter gives smoother cost function, but its flatter minimum may 
complicate discontinuity localization. } \label{fig:synthetic}
\end{figure*}

\begin{figure*}[h]
\centering
\begin{tabular}{ccc}
\includegraphics[width=1.8in,trim={5cm 0.6cm 2cm 4cm},clip]{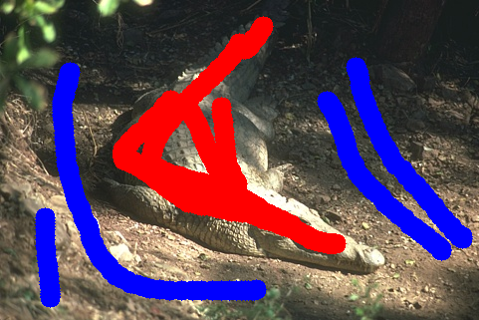}   &
\includegraphics[width=1.8in,trim={5cm 0.6cm 2cm 4cm},clip]{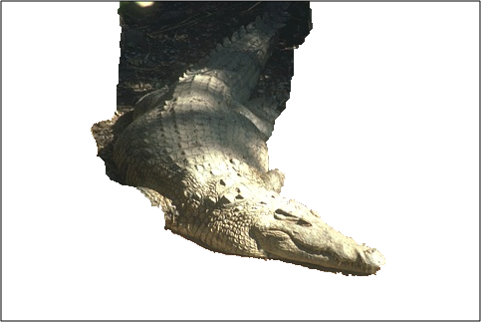}  &
\includegraphics[width=1.8in,trim={5cm 0.6cm 2cm 4cm},clip]{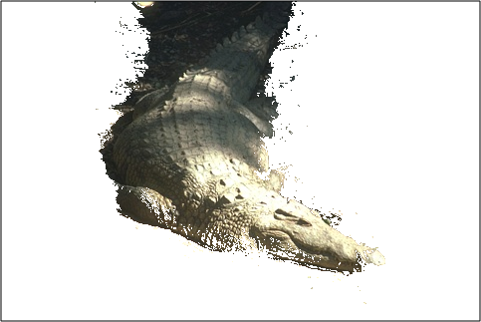}   \\
(a) image + seeds  & (b) grid CRF \cite{BJ:01} & (c)  dense CRF  \cite{koltun:NIPS11}
\end{tabular}
\caption{{\em Real "shallow" segmentation example} for sparse (b) and dense (c) CRF (Potts) models for image with seeds (a).
Sparse Potts gives smoother segment boundary with better edge alignment, while dense CRF inference often gives noisy boundary.} \label{fig:shallow}
\end{figure*}

Robust pairwise Potts model and its binary version (Ising model) are used in many application 
such as stereo, reconstruction, and segmentation. One can define this model as a cost functional over 
integer-valued labeling $S:=(S_p \in Z^+ \,|\,p\in\Omega)$ of image pixels $p\in\Omega$ as follows 
\begin{equation} \label{eq:potts}
E_P (S) \;\;=\;\; \sum_{pq\in\N} w_{pq} \cdot [S_p \neq S_q]
\end{equation}
where $\N$ is a given neighborhood system, $w_{pq}$ is a discontinuity penalty between
neighboring pixels $\{p,q\}$, and $[\cdot]$ is {\em Iverson bracket}. 
The nearest-neighbor version over $k$-connected grid $\N_k$,
as well as its popular variational analogues, \eg geodesic active contours \cite{Sapiro:97}, convex relaxations 
\cite{pock:CVPR09,chambolle2011first}, or continuous max-flow \cite{Yuan:CVPR2010}, are particularly well-researched. It is common to use contrast-weighted discontinuity 
penalties \cite{BVZ:PAMI01,BJ:01} between the neighboring points, as emphasized by the condition $\{pq\}\in\N_k$ below
\begin{equation} \label{eq:NNweights}
w_{pq} \;\;\;\; = \;\;\;\; \lambda\cdot\exp{\frac{-||I_p-I_q||^2}{2\sigma^2}}\;\;\cdot\;\;[\{pq\}\in\N_k].
\end{equation}
Nearest neighbor Potts models minimize the contrast-weighted length of the segmentation boundary preferring 
shorter perimeter aligned with image edges, \eg see Fig.~\ref{fig:shallow}(b).
The popularity of this model can be explained by generality, robustness, well-established foundations in geometry, 
and a large number of efficient discrete or continuous solvers that guarantee global optimum
in binary problems \cite{BJ:01} or some quality bound in multi-label settings, \eg $\alpha$-expansion~\cite{BVZ:PAMI01}.

Dense CRF \cite{koltun:NIPS11} is a Potts model where pairwise interactions are
active over significantly bigger neighborhoods defined by a Gaussian kernel with a relatively large bandwidth $\Delta$ 
over pixel locations
\begin{equation} \label{eq:NNweightsColt}
w_{pq} \;\; = \;\; \lambda\cdot\exp{\frac{-||I_p-I_q||^2}{\sigma^2}}\;\cdot\;\exp{\frac{-\|p-q\|^2}{\Delta^2}}.
\end{equation}
Its use in shallow vision is limited as it often produces noisy boundaries \cite{koltun:NIPS11}, 
see also Fig.~\ref{fig:shallow}(c). Also, global optimization methods mentioned above
do not scale to dense neighborhoods. Yet, dense CRF model is popular in the context of CNNs 
where it can be used as a differentiable regularization layer \cite{zheng2015conditional,schwing2015fully}. Larger bandwidth yields 
smoother objective \eqref{eq:potts}, see Fig. \ref{fig:synthetic}(c), amenable to gradient descent or 
other local linearization methods like mean-field inference that are easy to parallelize. Note that existing
efficient inference methods for dense CRF require {\em  bilateral filtering} \cite{koltun:NIPS11}, 
which is restricted to Gaussian weights as in \eqref{eq:NNweightsColt}. 
This is in contrast to global Potts solvers, \eg $\alpha$-expansion, that can use arbitrary weights, 
but become inefficient for dense neighborhoods. 

Noisier dense CRF results, \eg in Fig.~\ref{fig:shallow}(c), imply weaker regularization. Indeed, 
as discussed in \cite{veksler2018efficient}, for larger neighborhoods the Potts model gets closer 
to {\em cardinality potentials}.
Bandwidth $\Delta$ in \eqref{eq:NNweightsColt} is a resolution scale at which the model sees the segmentation boundary. 
Weaker regularization in dense CRF may preserve some thin structures smoothed out 
by fine-resolution boundary regularizers, \eg nearest-neighbor Potts.
However, this is essentially the same ``noise preservation'' effect shown in Fig.~\ref{fig:shallow}(c). 
For consistency, the rest of the paper refers to the nearest-neighbor Potts model as {\em grid CRF}, 
and large-neighborhood Potts as {\em dense CRF}.

\subsection{Summary of Contributions} 
\label{sec:contributions}

Any motivation for standard regularization models in shallow image segmentation, as in the previous section,
directly translates into their motivation as regularized loss functions in weakly supervised CNN segmentation \cite{NCloss:CVPR18,Rloss:ECCV18}.
The main issue is how to optimize these losses. Standard training techniques based on gradient descent may not be 
appropriate for many powerful regularization models, which may have many local minima.
Below is the list of our main contributions:
\begin{itemize}
\item As an alternative to gradient descent (GD), we propose a splitting technique, \emph{alternating direction method} (ADM)\footnote{Standard ADM\textit{\textbf{M}}~\cite{Boyd2011} casts a problem $\min_{x} f(x) + g(x)$ into $\min_{x,y}\max_\lambda f(x) + g(y) + \lambda^\mathrm{T}(x-y) + \rho\|x-y\|^2$ and alternates 
updates over $x$, $y$ and $\lambda$ optimizing $f$ and $g$ in parallel. Our ADM uses a different form of splitting and can be seen
as a {\em penalty method}, see Sec.~\ref{sec:ADM}.}, for minimizing {\em regularized losses} 
during network training. ADM can directly employ efficient regularization solvers known in shallow segmentation. 
\item Compared to GD, our ADM approach with $\alpha$-expansion solver significantly improves optimization quality for 
the {\em grid CRF} (nearest-neighbor Potts) loss in weakly supervised CNN segmentation. While each iteration of ADM is slower than GD,
the loss function decreases at a significantly larger rate with ADM. In one step it can reach lower loss values than those where GD converges. Grid CRF has never been investigated as loss for CNN segmentation and is largely overlooked.
\item The training quality with grid CRF loss achieves the-state-of-the-art in weakly supervised CNN segmentation.
We compare dense CRF and grid CRF losses. 
\end{itemize}

Our results may inspire more research on regularized segmentation losses and their optimization.

\section{ADM for Regularized Loss Optimization}
\label{sec:ADM}

Assume there is a dataset of pairs of images and partial ground truth labelings. 
For simplicity of notation, we implicitly assume summation over all pairs in 
dataset for all expressions of loss functions.
For each pixel $p \in \Omega$ of each image $I$ there is  an associated
color or intensity $I_p$ of that pixel. The labeling $Y=(Y_p|p \in \Omega_{\cal L})$ 
where $\Omega_{\cal L} \subset \Omega$ is a set of labeled pixels,
each $Y_p \in \{0,1\}^K$ is a one-hot distribution and $K$ is the number of labels.
We consider a regularized loss for network $\phi_\theta$ of the following form
\begin{equation}
\label{CNN-regularized-loss}
\ell (S_\theta,Y) \;\;+ \;\;\lambda\cdot  R(S_\theta) \;\;\;\; \to \;\;\;\; \min_\theta
\end{equation}
where $S_\theta \in [0,1]^{|\Omega|\times K}$ is a $K$-way softmax segmentation 
generated by the network $S_\theta := \phi_\theta(I)$,
and $R(\cdot)$ is a regularization term, \eg \emph{relaxed} sparse Potts or dense CRF, 
and $\ell (\cdot,\cdot)$ is a partial ground-truth loss, for instance: 
$$
\ell (S_\theta,Y) \; = \sum_{p\in\Omega_{\cal L}} {H}(Y_p,S_{p, \theta}), 
$$
where $H(Y_p,S_{p, \theta})=-\sum_kY_p^k \log S_{p, \theta}^k$ 
is the cross entropy between  predicted segmentation 
$S_{p, \theta}$ (a row of matrix $S_{\theta}$ corresponding to pixel $p$) 
and ground truth labeling $Y_p$. 

We present a general alternating direction method (ADM) to optimize neural network regularized losses 
of the general form in \eqref{CNN-regularized-loss} using the following splitting of the problem:
\begin{equation}
	\label{}
	\begin{aligned}
	& \underset{\theta,X}{\text{minimize}} & & \ell (S_\theta,Y) \; + \; \lambda R(X)  \\
	& \text{subject to} & & \sum_{p\in\Omega_{{\cal U}}} D(X_p|S_{p,\theta}) = 0, \\
	& & & X_p = Y_p & \forall p \in \Omega_{\cal L}
	\end{aligned}
\end{equation}
where we introduced one-hot distributions $X_p \in \{ 0,1 \}^K$ and 
some divergence measure $D$, \eg the Kullback-Leibler divergence.
$R(X)$ can now be a \emph{discrete} classic MRF regularization, \eg \eqref{eq:potts}.
This equates to the following Lagrangian 
\begin{equation}
\label{Alternating-direction-KL}
\begin{aligned}
& \min_{\theta,X}\max_\gamma & & \ell (S_\theta,Y) + \lambda R(X) + \gamma \!\sum_{p\in\Omega_{{\cal U}}} \!D(X_p|S_{p,\theta})\\
& \text{subject to} & & X_p = Y_p \;\;\; \forall p \in \Omega_{\cal L}.
\end{aligned}
\end{equation}

We alternate optimization over $X$ and $\theta$ in \eqref{Alternating-direction-KL}.
The maximization over $\gamma$ increases its value at every update resulting
in a variant of simulated annealing. We have experimented with variable multiplier $\gamma$
but found no advantage compared to fixed $\gamma$.
So, we fix $\gamma$ and do not optimize for it.
In summary, instead of optimizing the regularization term 
with gradient descent, our approach splits 
regularized-loss problem \eqref{CNN-regularized-loss} into two sub-problems. 
We replace the softmax outputs $S_{p,\theta}$ in the regularization term by latent discrete variables $X_p$ and ensure consistency between both variables 
(\ie, $S_\theta$ and $X$) by minimizing divergence $D$. 

This is similar conceptually to the general principles of 
ADMM~\cite{Boyd2011,NIPS2014_5612}. Our ADM splitting accommodates the use of powerful and well-established 
discrete solvers for the regularization loss. As we show in Sec.~\ref{sec:experiments}, the popular $\alpha$-expansion solver \cite{BVZ:PAMI01} significantly improves 
optimization of grid CRF losses yielding state-of-the-art training quality. Such efficient solvers guarantee global optimum in binary problems \cite{BJ:01} or a quality bound in multi-label case~\cite{BVZ:PAMI01}.

Our discrete-continuous ADM method\footnote{Combining continuous and discrete sub-problem solvers is not uncommon 
in tailored ADMM-inspired splitting algorithms \cite{Torr:BMVC14,dtADMM:ICIP16,Ismail:CVPR17,Ismail:miccai17,Manu:CVPR18}.} 
alternates two steps, 
each decreasing \eqref{Alternating-direction-KL}, until convergence. 
Given fixed discrete latent variables $X_p$ computed at the previous iteration, 
the first step learns the network parameters $\theta$ by minimizing 
the following loss via standard back-propagation and a variant of 
stochastic gradient descent (SGD): 
\begin{equation}
\label{SGD}
\underset{\theta}{\text{minimize}} \;\; \ell (S_\theta,Y) \; + \; \gamma \sum_{p\in\Omega_{{\cal U}}} D(X_p|S_{p,\theta})
\end{equation}

The second step fixes the network output $S_\theta$ and finds the next 
latent binary variables $X$ by minimizing the following 
objective over $X$ via any suitable discrete solver: 
\begin{equation}
\label{Alpha-expansion}
\begin{aligned}
& \underset{{X \in \{ 0,1 \}^{|\Omega|\times K}}}{\text{minimize}} & &\lambda R(X) \; +  \; \gamma \sum_{p\in\Omega_{{\cal U}}} D(X_p|S_{p,\theta})\\
& \text{\;\;\,subject to} & & X_p = Y_p \;\;\;\;\; \forall p \in \Omega_{\cal L}.
\end{aligned}
\end{equation}
Because $X_p$ is a discrete variable with only $K$ possible values, 
the second term in \eqref{Alpha-expansion} is a basic unary term. 
Similarly, the equality constraints could be implemented as unary terms 
using prohibitive values of unary potentials. Unary terms are simplest
possible energy potentials that can be handled by any
general discrete solver. On the other hand, the regularization term $R(X)$ 
usually involves interactions of two or more variables introducing
new properties of solution together with optimization complexity. 
In case of grid CRF one can use graph cut \cite{BJ:01}, 
$\alpha$-expansion \cite{BVZ:PAMI01}, QPBO \cite{qpbo,qpbo-i}, TRWS \cite{trws}, LBP \cite{lbp}, LSA-TR \cite{lsa-tr} \etc.

In summary, our approach alternates the two steps described above. 
For each minibatch we compute network prediction, 
then compute hidden variables $X$ optimizing \eqref{Alpha-expansion},
then compute gradients of loss \eqref{SGD} and update the parameters of the
network using a variant of SGD. The outline of our ADM scheme is shown in Alg.~\ref{alg:adm}.

\begin{algorithm}
\begin{algorithmic} 
    \REQUIRE sequence of minibatches
    \STATE $i \leftarrow 0$;
    \STATE initialize network parameters $\theta^{(0)}$;
    \FOR{each minibatch $B$}
	    \FOR{each image-labeling pair $(I,Y)\in B$}
	        \STATE Compute segmentation prediction $S_{\theta} \leftarrow \phi_{\theta^{(i)}}(I)$;
	        \STATE Solve energy \eqref{Alpha-expansion} for $X$ with \eg $\alpha$-expansion;
	        \STATE Compute gradients $g$ w.r.t. parameters $\theta$ of \eqref{SGD};
	    \ENDFOR
	    \STATE Compute average over the batch $g^{(i)} \leftarrow \frac1{|B|}\sum g$;
	    \STATE Update network parameters $\theta^{(i+1)}$ using gradient $g^{(i)}$;
	    \STATE $i \leftarrow i + 1$;
    \ENDFOR
\end{algorithmic}
\caption{ADM for regularized loss~\eqref{CNN-regularized-loss}.}
\label{alg:adm}
\end{algorithm}
 


\section{Experimental Results}
\label{sec:experiments}

We conduct experiments for weakly supervised CNN segmentation with scribbles as supervision~\cite{scribblesup}. The focus is on regularized loss approaches~\cite{NCloss:CVPR18,Rloss:ECCV18} yet we also compare our results to proposal generation based method, \eg ScribbleSup \cite{scribblesup}. We test both the grid CRF and dense CRF as regularized losses. Such regularized loss can be optimized by stochastic gradient descent (GD) or alternative direction method (ADM), as discussed in Sec.~\ref{sec:ADM}. We compare three training schemes, namely dense CRF with GD \cite{Rloss:ECCV18}, grid CRF with GD and grid CRF with ADM for weakly supervised CNN segmentation.

Before comparing segmentations, in Sec.~\ref{sec:ADMvsGD} we test if using ADM gives better regularization losses than that using standard GD. Our plots of training losses (CRF energy) vs training iterations show how fast the losses converge when minimized by ADM or GD. Our experiment confirms that first order approach like GD leads to a poor local minimum for the grid CRF loss. There are clear improvements of ADM over GD for minimization of the grid CRF loss. In Sec.~\ref{sec:mIOU}, rather than comparing in terms of optimization, we compare ADM and GD in terms of segmentation quality. We report both mIOU (mean intersection over union) and accuracy in particular for boundary regions. In Sec.~\ref{sec:shortenedscribbles}, we also study these variants of regularized loss method in a more challenging setting of shorter scribbles~\cite{scribblesup} or clicks in the extreme case. With ADM as the optimizer, our approach of grid CRF regularized loss compares favorably to dense CRF based approach \cite{Rloss:ECCV18}. 

\paragraph{Dataset and implementation details} Following recent work \cite{deeplab,scribblesup,kolesnikov2016seed,NCloss:CVPR18} on CNN semantic segmentation, we report our results on PASCAL VOC 2012 segmentation dataset. We train with scribbles from~\cite{scribblesup} on the augmented datasets of 10,582 images and test on the \textit{val} set of 1,449 images. We report mIOU (mean intersection over union) and pixel-wise accuracy. In particular, we are interested in how good is the segmentation in the boundary region. So we compute accuracy for those pixels close to the boundary, for example within 8 or 16 pixels from semantic boundaries. Besides mIOU and accuracy, we also measure the regularization losses, \ie the grid CRF. Our implementation is based on DeepLabv2\footnote{https://bitbucket.org/aquariusjay/deeplab-public-ver2} and we show results on different networks including deeplab-largeFOV, deeplab-msc-largeFOV, deeplab-vgg16 and resnet-101. We do not apply any post-processing to network output segmentation.

The networks are trained in two phases. Firstly, we train the network to minimize partial cross entropy (pCE) loss w.r.t scribbles. Then we train with a grid CRF or dense CRF regularization term. To implement gradient descent for the discrete grid CRF loss, we first take its quadratic relaxation,
\begin{equation}
\langle\mathbf{1},S_{p,\theta}\rangle+\langle\mathbf{1},S_{q,\theta}\rangle-2\langle S_{p,\theta},S_{q,\theta}\rangle.
\end{equation}
where $S_{p,\theta},S_{q,\theta}\in [0,1]^K$ and $\langle\cdot,\cdot\rangle$ is the dot product. Then we differentiate \wrt $S_{\theta}$ during back-propagation. While there are ways, \eg \cite{grady2011,chambolle2012convex}, to relax discrete Pott's model, we focus on this simple and standard relaxation~\cite{yuille2010belief,koltun:NIPS11,Rloss:ECCV18}.

For our proposed ADM algorithm, which requires inference in the grid CRF, we use a public implementation of $\alpha$-expansion\footnote{http://mouse.cs.uwaterloo.ca/code/gco-v3.0.zip}. The CRF inference and loss are implemented and integrated as Caffe \cite{caffe} layers. We run $\alpha$-expansion for five iterations, which in most cases gives convergence. Our dense CRF loss does not include the Gaussian kernel on locations $XY$, since ignoring this term does not change the mIOU measure \cite{koltun:NIPS11}. The bandwidth for the dense Gaussian kernel on $RGBXY$ is validated to give the best mIOU. For the grid CRF, the kernel bandwidth selection in \eqref{eq:NNweights} follows standard Boykov-Jolly \cite{BJ:01}
$$\sigma^2 \;\; = \;\; \frac1{|N|} \sum_{pq \in N}{ \|I_p - I_q\|^2}.$$

In general, our ADM optimization for regularized loss is slower than GD due to the inference of grid CRF. However, for inference algorithms, \eg $\alpha$-expansion, that cannot be easily parallelized, we utilize simple multi-core parallelization for all images in a batch to accelerate training. Note that we do not use CRF inference during testing.

\subsection{Loss Minimization}
\label{sec:ADMvsGD}

\begin{figure}[h]
    \centering
    \begin{tabular}{ccc}
    \includegraphics[width=0.45\textwidth]{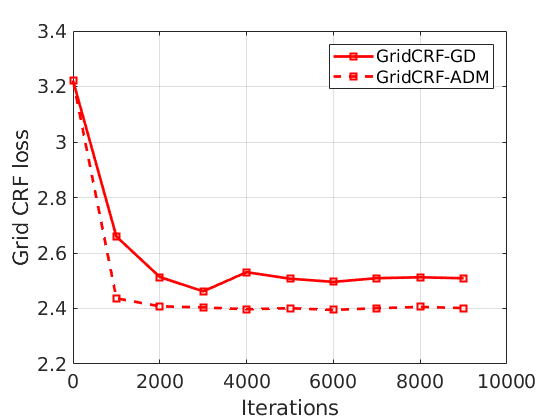} 
     \\
   \end{tabular}
   \caption{Training progress of ADM and gradient descend (GD) on Deeplab-MSc-largeFOV. Our ADM for the grid CRF loss with $\alpha$-expansion significantly improves convergence and achieves lower training loss. For example, first 1,000 iterations of ADM give grid CRF loss lower than GD's best result.
   }
   \label{fig:iterations}
\end{figure}

In this section, we show that for grid CRF losses the ADM approach employing $\alpha$-expansion~\cite{BVZ:PAMI01}, a powerful discrete optimization method, outperforms common gradient descend methods for regularized losses \cite{NCloss:CVPR18,Rloss:ECCV18} in terms of finding a lower minimum of regularization loss.  Tab.~\ref{tb:gridCRFloss} shows the grid CRF losses on both training and validation sets for different network architectures. Fig.~\ref{fig:iterations}(a) shows the evolution of the grid CRF loss over the number of iterations of training. ADM requires fewer iterations to achieve the same CRF loss. The networks trained using ADM scheme give lower CRF losses for both training and validation sets.

\begin{table}[bt]
\renewcommand{\arraystretch}{1.3}
\centering
\setlength\tabcolsep{5pt}
\begin{tabular}{| l | c | c | c | c|}
\hline
\multirow{ 2}{*}{network}    &       \multicolumn{2}{|c|}{training set$^\dag$}    &  \multicolumn{2}{|c|}{validation set}      \\ \cline{2-5}
&        GD             &        ADM        & GD       & ADM  \\ \hline
   Deeplab-LargeFOV    &       2.52              &     2.41        & 2.51  &  2.33 \\ \hline
   Deeplab-MSc-largeFOV    &        2.51              &     2.40        & 2.49   &  2.33     \\ \hline
    Deeplab-VGG16   &       2.37             &        2.10        &  2.42    &     2.14 \\ \hline
     Resnet-101    &        2.66             &        2.49       & 2.61    &   2.42  \\ \hline
\end{tabular}
\vspace{1ex}
\caption{ADM gives better grid CRF losses than gradient descend (GD). \dag We randomly selected 1,000 training examples.}\label{tb:gridCRFloss}
\end{table}


\begin{figure*}[t]
\setlength{\tabcolsep}{1pt}
\newcommand{\oneseven}{0.156\textwidth}
\centering
\begin{tabular}{@{}ccccccc}
\includegraphics[width=\oneseven,trim=0 0 0 0,clip]{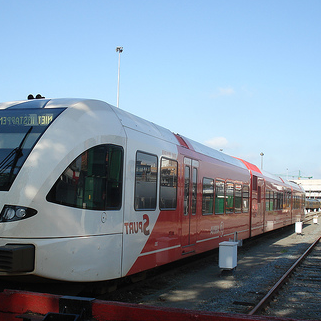}& 
\includegraphics[width=\oneseven,trim=0 0 0 0,clip]{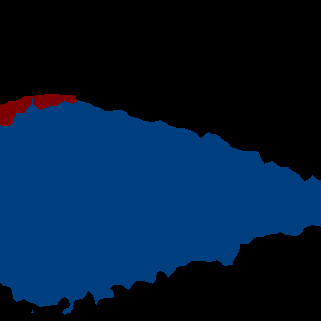}&
\includegraphics[width=\oneseven,trim=2.75cm 1.35cm 4.2cm 1.51cm,clip]{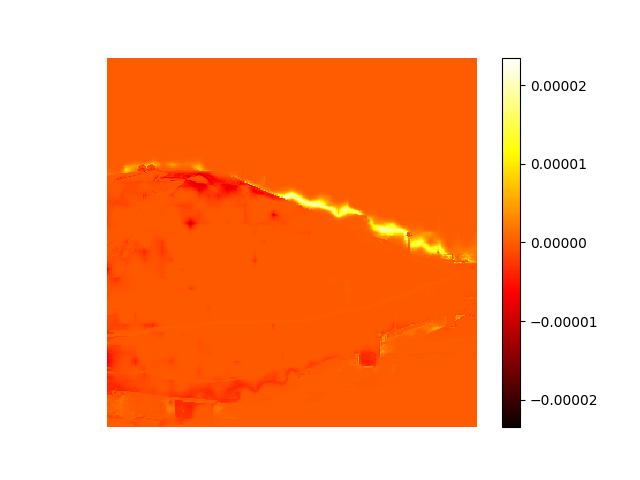} & 
\includegraphics[width=\oneseven,trim=2.75cm 1.35cm 4.2cm 1.51cm,clip]{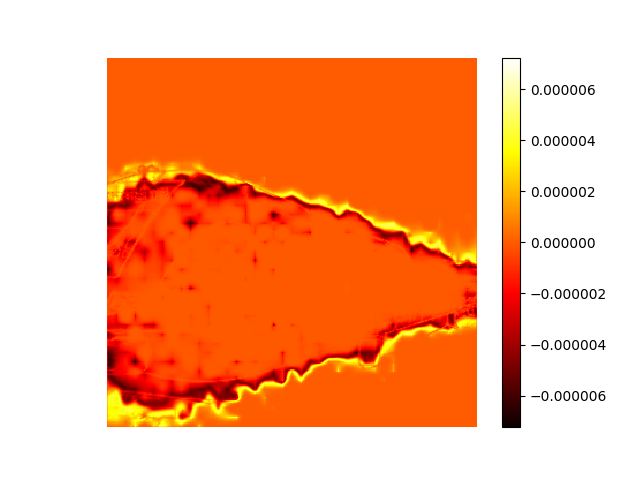} & 
\includegraphics[width=\oneseven,trim=2.75cm 1.35cm 4.2cm 1.51cm,clip]{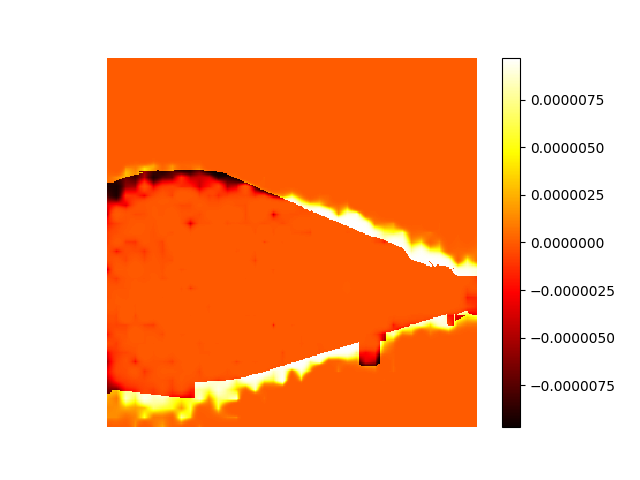}  \\
\includegraphics[width=\oneseven,clip,trim=0 0 0 5cm]{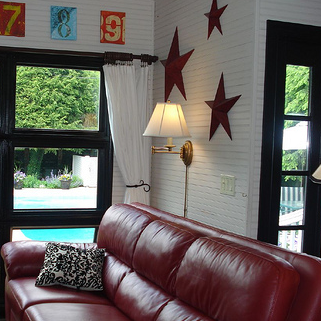} & 
\includegraphics[width=\oneseven,clip,trim=0 0 0 5cm]{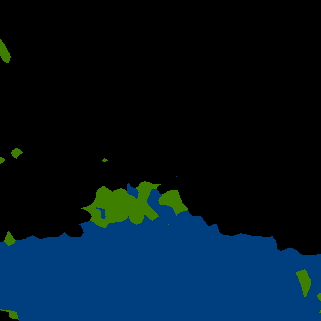} &
\includegraphics[width=\oneseven,clip,trim=0 0 0 5cm]{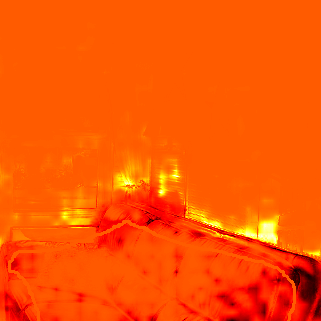} & 
\includegraphics[width=\oneseven,clip,trim=0 0 0 5cm]{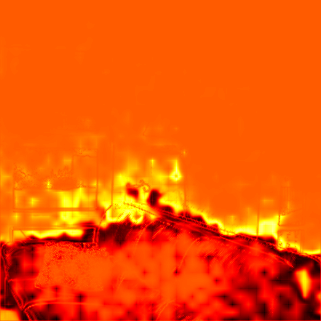} & 
\includegraphics[width=\oneseven,clip,trim=0 0 0 5cm]{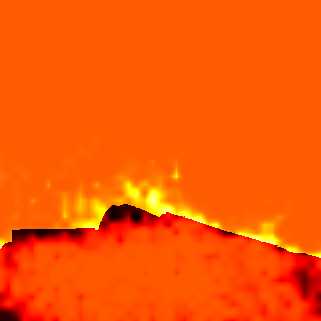}  \\
\includegraphics[width=\oneseven,clip,trim=0 6cm 0 0]{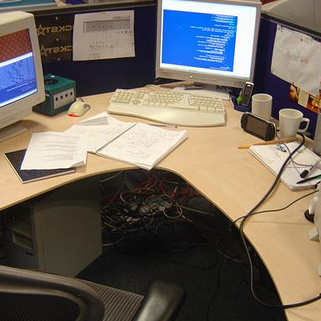}   & 
\includegraphics[width=\oneseven,clip,trim=0 6cm 0 0]{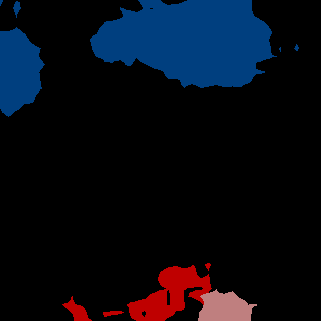} &
\includegraphics[width=\oneseven,clip,trim=0 6cm 0 0]{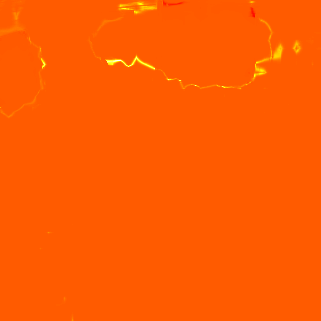} & 
\includegraphics[width=\oneseven,clip,trim=0 6cm 0 0]{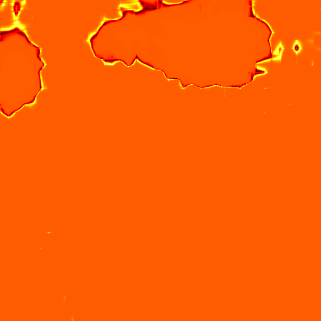} & 
\includegraphics[width=\oneseven,clip,trim=0 6cm 0 0]{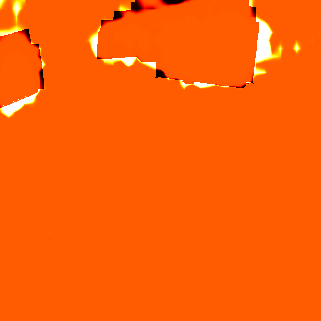}   \\
\includegraphics[width=\oneseven,clip,trim=0 0 0 5cm]{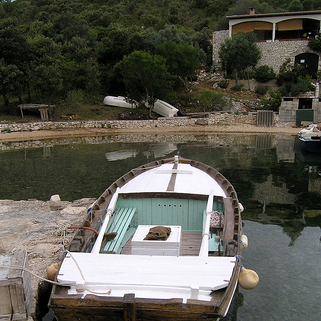}   & 
\includegraphics[width=\oneseven,clip,trim=0 0 0 5cm]{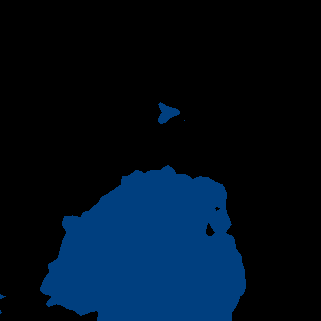} &
\includegraphics[width=\oneseven,clip,trim=0 0 0 5cm]{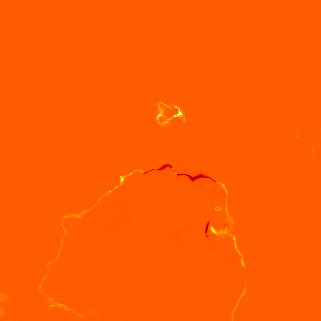} & 
\includegraphics[width=\oneseven,clip,trim=0 0 0 5cm]{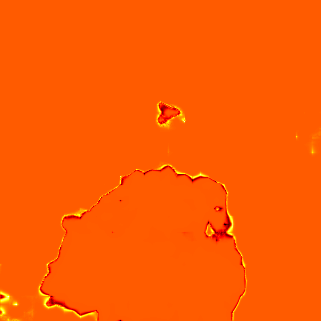} & 
\includegraphics[width=\oneseven,clip,trim=0 0 0 5cm]{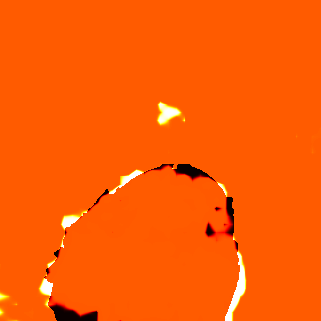} \\
\includegraphics[width=\oneseven,clip,trim=0 0 0 2cm]{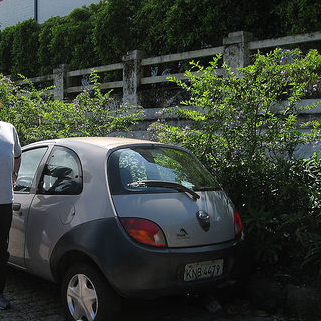}   & 
\includegraphics[width=\oneseven,clip,trim=0 0 0 2cm]{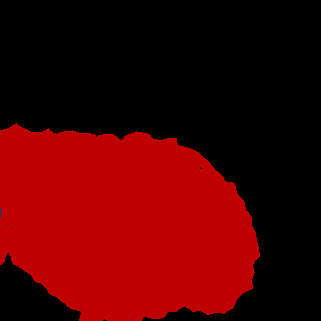} &
\includegraphics[width=\oneseven,clip,trim=0 0 0 2cm]{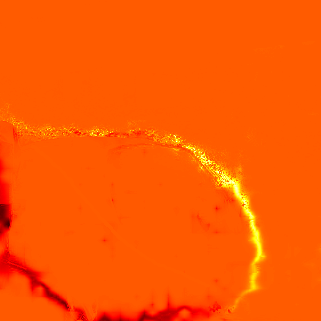} & 
\includegraphics[width=\oneseven,clip,trim=0 0 0 2cm]{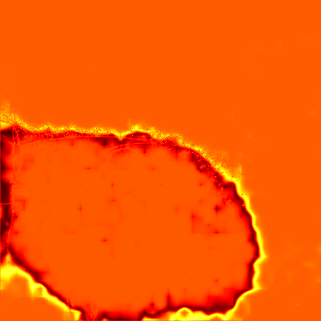} & 
\includegraphics[width=\oneseven,clip,trim=0 0 0 2cm]{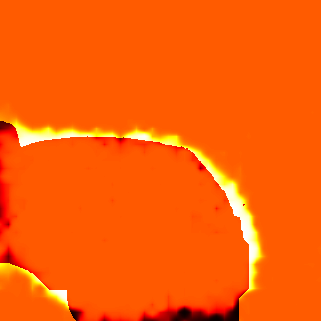} \\
\includegraphics[width=\oneseven]{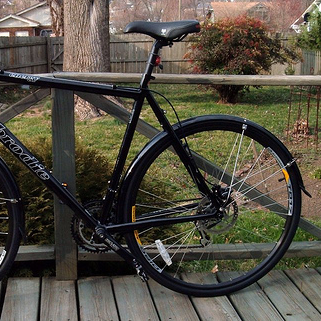}   & 
\includegraphics[width=\oneseven]{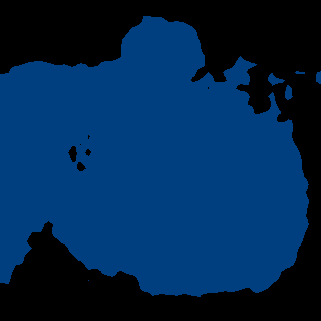} &
\includegraphics[width=\oneseven]{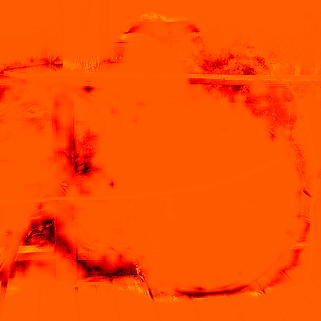} & 
\includegraphics[width=\oneseven]{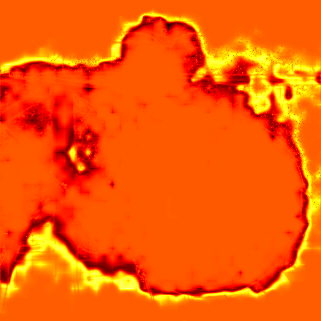} & 
\includegraphics[width=\oneseven]{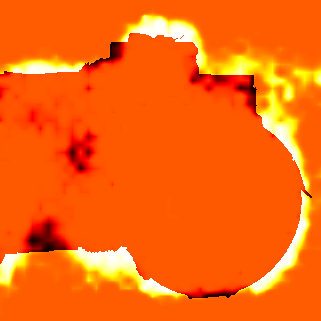}  \\
(a) input & (b) prediction &(c) Dense GD\cite{Rloss:ECCV18} & (d) Grid GD & (e) Grid ADM
\end{tabular}
\caption{The gradients with respect to scores of the deeplab\_largeFOV network with the dense CRF (c) and grid CRF (d~and~e for using either the plain stochastic gradient descent or our ADM scheme). Latent segmentation in ADM with the grid CRF loss produces gradients more directly pointing to a good solution (e). Note, the object boundaries are more prominent in (e).} \label{fig:gradients}
\end{figure*}

\vspace{-2ex}
\paragraph{The gradients} with respect to the soft-max layer's input of the network are visualized in Fig.~\ref{fig:gradients}. Clearly, our ADM approach with the grid CRF enforces better edge alignment. Despite different formulations of regularized losses and their optimization, the gradients of either~\eqref{CNN-regularized-loss} or~\eqref{SGD} \wrt network output $S_\theta$ are the driving force for training. In most of the cases, GD produces significant gradient values only in the vicinity of the current model prediction boundary as in Fig.~\ref{fig:gradients}(c,d). If the actual object boundary is sufficiently distant the gradient methods fail to detect it due to the sparsity of the grid CRF model, see Fig.~\ref{fig:synthetic} for an illustrative ``toy'' example. On the other hand, the ADM method is able to predict a good latent segmentation allowing gradients leading to a good solution more effectively, see Fig.~\ref{fig:gradients}(e). 

Thus, in the context of grid CRFs, the ADM approach coupled with $\alpha$-expansion shows drastic improvement in the optimization quality. In the next section, we further compare ADM with GD to see which gives better segmentation.

\subsection{Segmentation Quality}
\label{sec:mIOU}

The quantitative measures for segmentation by different methods are summarized in Tab.~\ref{tab:mainresult} and Tab.~\ref{tab:accuracy}. The mIOU and segmentation accuracy on the \textit{val} set of PASCAL 2012~\cite{pascal-voc-2012} are reported for various networks. The supervision is scribbles~\cite{scribblesup}. The quality of weakly supervised segmentation is bounded by that with full supervision and we are interested in the gap for different weakly supervised approaches. 

The baseline approach is to train the network using proposals generated by GrabCut style interactive segmentation with such scribbles. Besides the baseline (train w/ proposals), here we compare variants of regularized losses optimized by gradient descent or ADM. The regularized loss is comprised of the partial cross entropy (pCE) \wrt scribbles and grid/dense CRF. Other losses \eg normalized cut \cite{Shi2000,NCloss:CVPR18} may give better segmentation, but the focus is to compare gradient descent vs ADM optimization for the grid CRF.

It is common to apply dense CRF post-processing~\cite{deeplab} to the network's output during testing. However, for the sake of clear comparison, we show results without it.

As shown in Tab.~\ref{tab:mainresult}, all regularized approaches work better than the non-regularized approach that only minimizes the partial cross entropy. Also, the regularized loss approaches are much better than proposal generation based method since erroneous proposals may mislead training. 

Among regularized loss approaches, grid CRF with GD performs the worst due to the fact that a first-order method like gradient descent leads to the poor local minimum for the grid CRF in the context of energy minimization. Our ADM for the grid CRF gives much better segmentation competitive with the dense CRF with GD. The alternative grid CRF based method gives good quality segmentation approaching that for full supervision. Tab.~\ref{tab:accuracy} shows \emph{accuracy} of different methods for pixels close to the semantic boundaries. Such measure tells the quality of segmentation in boundary regions.

\begin{figure*}[h!]
\setlength{\tabcolsep}{1pt}
\newcommand{\oneseven}{0.16\textwidth}
\centering
\begin{tabular}{@{}ccccccc}
\includegraphics[width=\oneseven,trim=0 0 0 0,clip]{fig/exam/2009_000037.jpg}   & 
\includegraphics[width=\oneseven,trim=0 0 0 0,clip]{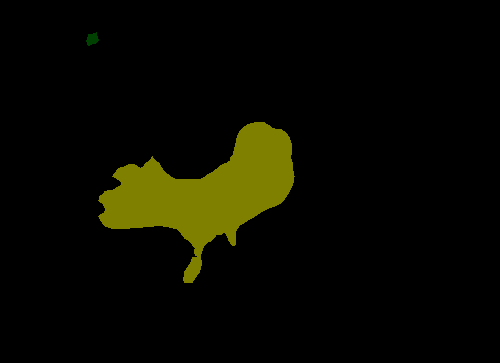} &
\includegraphics[width=\oneseven,trim=0 0 0 0,clip]{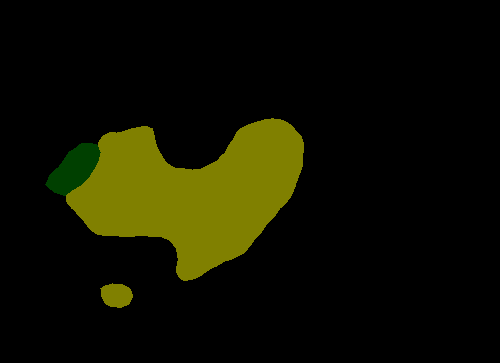} &
\includegraphics[width=\oneseven,trim=0 0 0 0,clip]{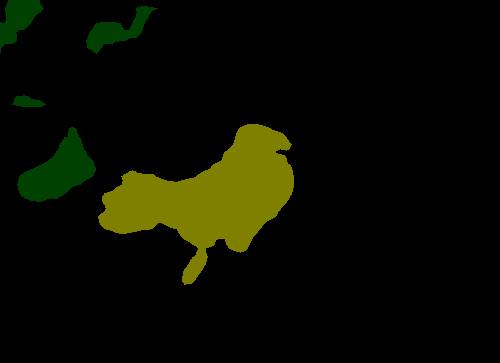} &
\includegraphics[width=\oneseven,trim=0 0 0 0,clip]{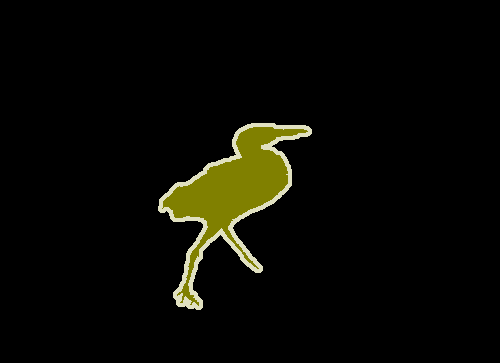} \\
\includegraphics[width=\oneseven,trim=0 0 0 0,clip]{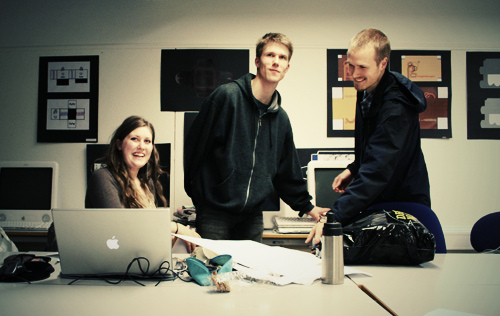}   & 
\includegraphics[width=\oneseven,trim=0 0 0 0,clip]{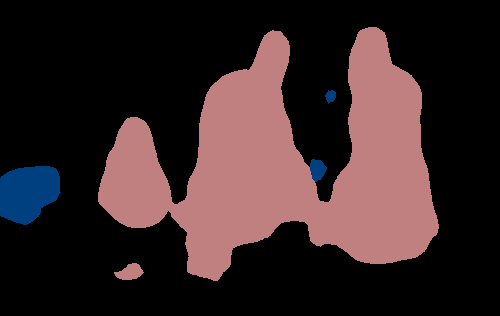} &
\includegraphics[width=\oneseven,trim=0 0 0 0,clip]{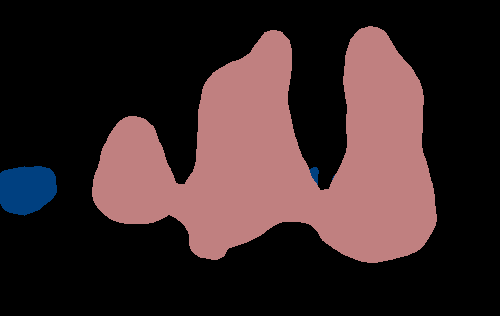} &
\includegraphics[width=\oneseven,trim=0 0 0 0,clip]{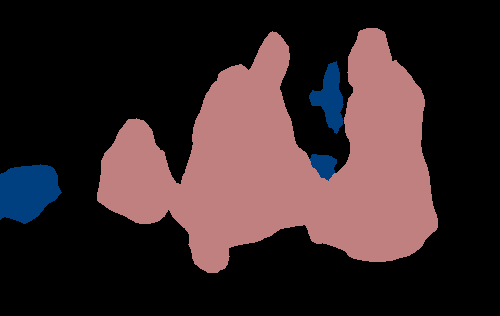} &
\includegraphics[width=\oneseven,trim=0 0 0 0,clip]{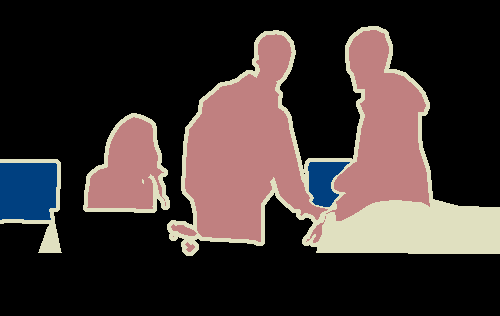} \\
\includegraphics[width=\oneseven,trim=0 0 0 0,clip]{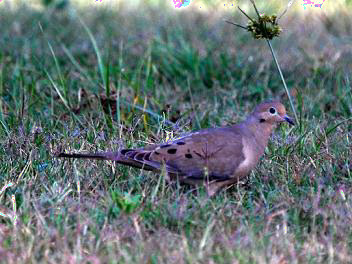}   & 
\includegraphics[width=\oneseven,trim=0 0 0 0,clip]{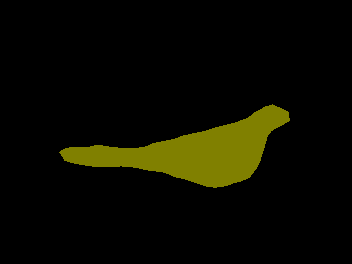} &
\includegraphics[width=\oneseven,trim=0 0 0 0,clip]{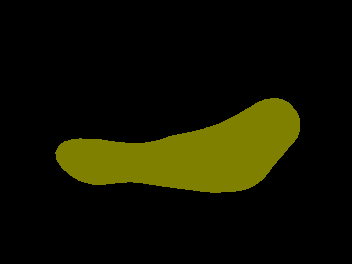} &
\includegraphics[width=\oneseven,trim=0 0 0 0,clip]{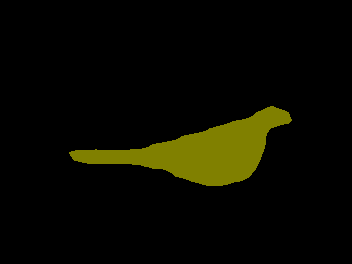} &
\includegraphics[width=\oneseven,trim=0 0 0 0,clip]{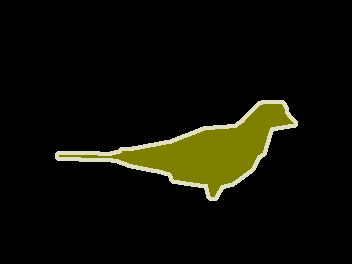} \\
\includegraphics[width=\oneseven,trim=0 0 0 0,clip]{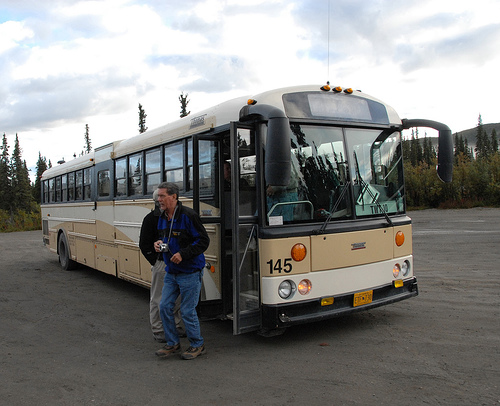}   & 
\includegraphics[width=\oneseven,trim=0 0 0 0,clip]{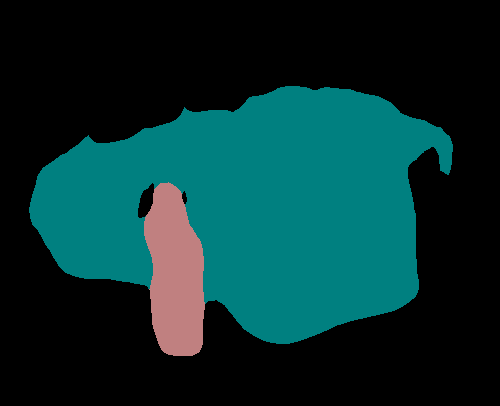} &
\includegraphics[width=\oneseven,trim=0 0 0 0,clip]{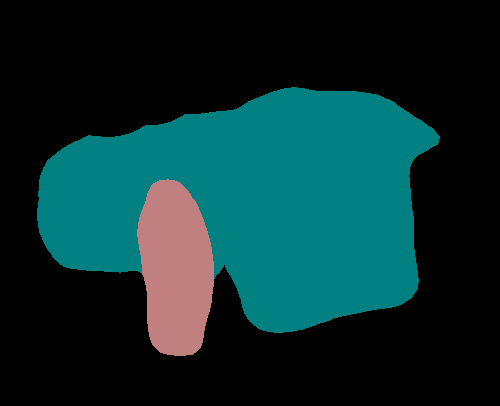} &
\includegraphics[width=\oneseven,trim=0 0 0 0,clip]{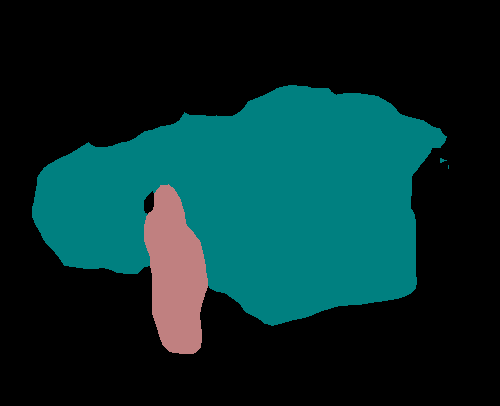} &
\includegraphics[width=\oneseven,trim=0 0 0 0,clip]{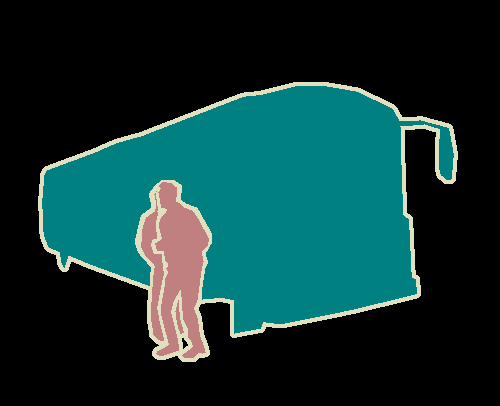} \\
\includegraphics[width=\oneseven,trim=0 0 0 0,clip]{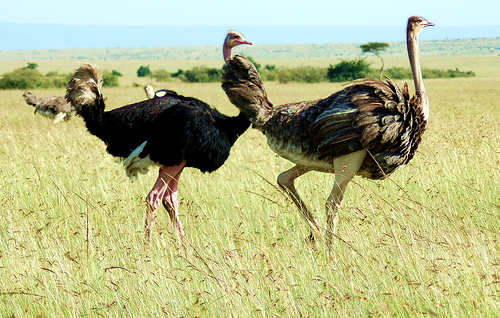}   & 
\includegraphics[width=\oneseven,trim=0 0 0 0,clip]{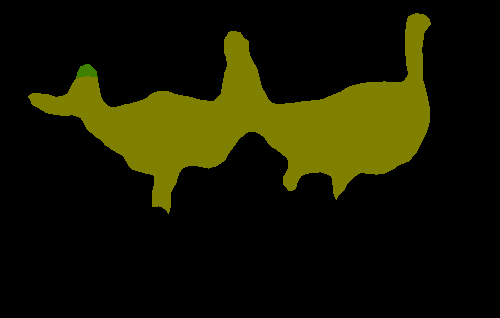} &
\includegraphics[width=\oneseven,trim=0 0 0 0,clip]{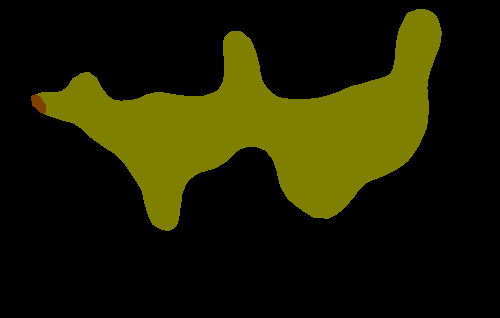} &
\includegraphics[width=\oneseven,trim=0 0 0 0,clip]{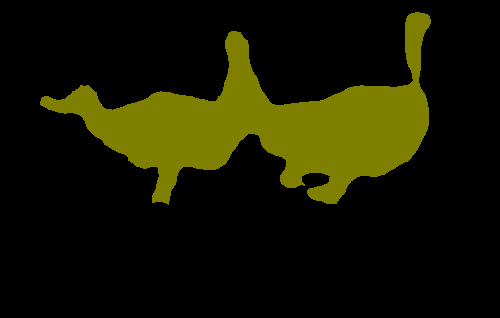} &
\includegraphics[width=\oneseven,trim=0 0 0 0,clip]{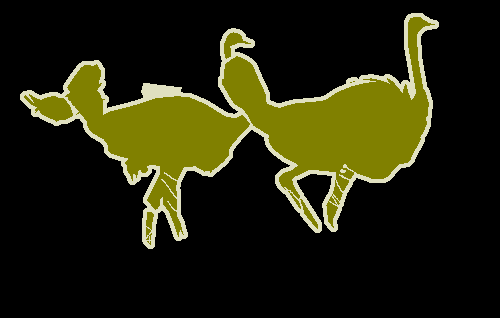} \\
\includegraphics[width=\oneseven,trim=0 0 0 0,clip]{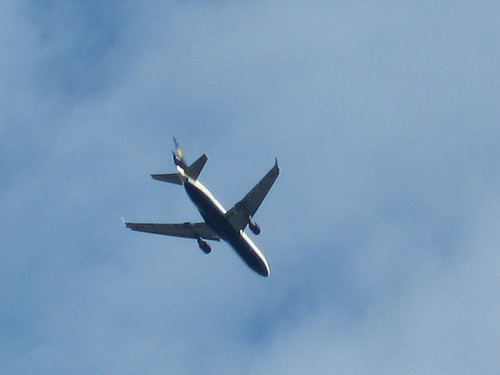}   & 
\includegraphics[width=\oneseven,trim=0 0 0 0,clip]{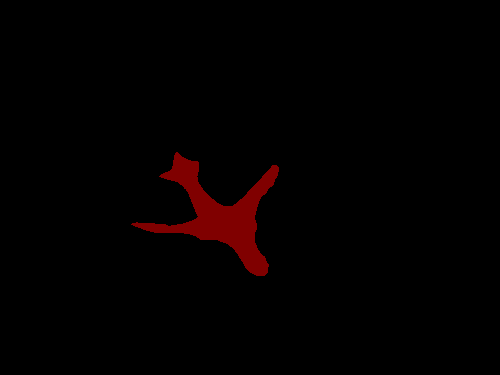} &
\includegraphics[width=\oneseven,trim=0 0 0 0,clip]{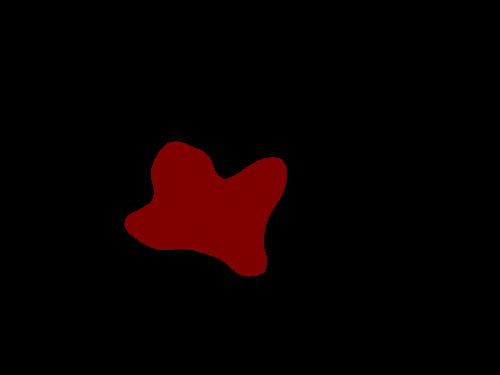} &
\includegraphics[width=\oneseven,trim=0 0 0 0,clip]{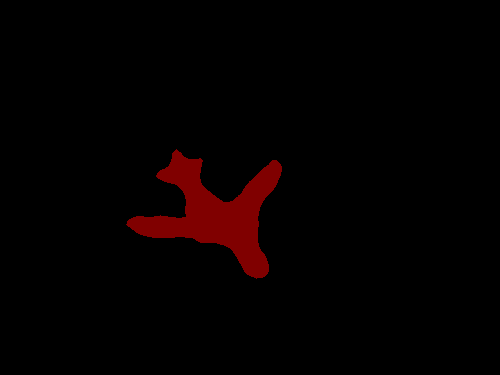} &
\includegraphics[width=\oneseven,trim=0 0 0 0,clip]{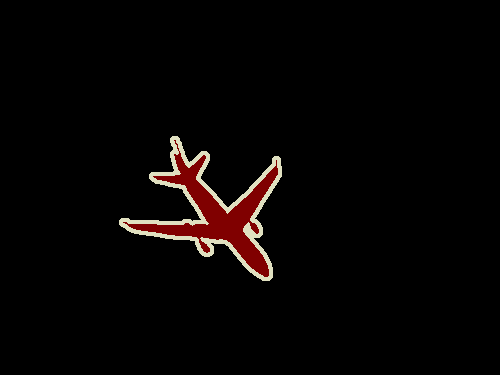} \\
(a) input & (b) Dense GD & (c) Grid GD & (d) Grid ADM & (e) ground truth
\end{tabular}
\caption{Example segmentations (Deeplab-MSc-largeFOV) by variants of regularized loss approaches. Gradient descent (GD) for grid CRF gives segmentation of poor boundary alignment though grid CRF is part of the regularized loss. ADM for grid CRF significantly improves edge alignment and compares favorably to dense CRF based method.} \label{fig:segexamples}
\end{figure*}

Fig.~\ref{fig:segexamples} shows a few qualitative segmentation results.

\begin{table*}[t]
\renewcommand{\arraystretch}{1.3}
\centering
\begin{tabular}{| l | c| c | c |c|  c |c|}
\hline
  & &\multicolumn{5}{|c|}{Weak supervision}  \\ \cline{3-7}
{Network}&  Full sup. &\multirow{ 2}{12mm}{train w/ proposals} &\multirow{ 2}{10mm}{\centering \centering pCE loss} & \multicolumn{1}{|c|}{+dense CRF loss} & \multicolumn{2}{|c|}{\cellcolor{gray!25}+grid CRF loss}      \\ \cline{5-7}
  &    &  & & GD \cite{Rloss:ECCV18} & {\cellcolor{gray!25} GD} &  {\cellcolor{gray!25} ADM} \\ \hline
Deeplab-largeFOV         & 63.0 & 54.8&55.8 & {\bf62.2}&{\cellcolor{gray!25} 60.4}&{\cellcolor{gray!25} 61.7} \\ \hline
Deeplab-MSc-largeFOV & 64.1  & 55.5 & 56 & {\bf63.1}  & {\cellcolor{gray!25} 61.2} & {\cellcolor{gray!25}62.9} \\ \hline
Deeplab-VGG16             & 68.8 &59.0& 60.4& 64.4  & {\cellcolor{gray!25} 63.3} & {\cellcolor{gray!25} {\bf65.2}} \\ \hline
ResNet-101                    & 75.6 &64.0& 69.5& {\bf72.9} & {\cellcolor{gray!25} 71.7 } & {\cellcolor{gray!25} 72.8} \\ \hline
\end{tabular}
\vspace{1ex}
\caption{Weakly supervised segmentation results for different choices of network architecture, regularized losses and optimization via gradient descent or ADM. We show mIOU on \textit{val} set of PASCAL 2012. ADM consistently improves over GD for different networks for grid CRF. Our grid CRF with ADM is competitive to previous state-of-the-art dense CRF (with GD) \cite{Rloss:ECCV18}.}
\label{tab:mainresult}
\end{table*}

\begin{table*}[h]
\renewcommand{\arraystretch}{1.3}
\centering
\begin{tabular}{l| l | c| c | c |c| c |c|}
\cline{2-8}
 & & &\multicolumn{5}{|c|}{Weak supervision}  \\ \cline{4-8}
& {Network}&  Full sup. &\multirow{ 2}{12mm}{train w/ proposals} &\multirow{ 2}{10mm}{\centering \centering pCE loss} & \multicolumn{1}{|c|}{+dense CRF loss} & \multicolumn{2}{|c|}{\cellcolor{gray!25}+grid CRF loss}      \\ \cline{6-8}
&  &    &  & & GD \cite{Rloss:ECCV18} & \cellcolor{gray!25}GD & \cellcolor{gray!25}ADM \\ \cline{2-8}
\multirow{ 3}{*}{\begin{sideways}\parbox{10ex}{\centering all\\pixels}\end{sideways}} &  Deeplab-MSc-largeFOV &  90.9 & 86.4  & 86.5 & {\bf90.6} &  \cellcolor{gray!25}89.9 & \cellcolor{gray!25}90.5 \\ \cline{2-8}
&  Deeplab-VGG16            &  91.6 & 88.6&88.9 & 91.1 & \cellcolor{gray!25}90.5 & \cellcolor{gray!25}{\bf91.3} \\ \cline{2-8}
&  ResNet-101                   & 94.5 & 90.2&92 &93.1  & \cellcolor{gray!25}92.9 & \cellcolor{gray!25}{\bf93.4} \\ \hline\hline
\multirow{ 3}{*}{\begin{sideways}\parbox{10ex}{\centering trimap\\16 pixels}\end{sideways}} &  Deeplab-MSc-largeFOV &  80.1 & 73.9  & 66.7 & {\bf77.8}  & \cellcolor{gray!25}74.8 & \cellcolor{gray!25}76.7 \\ \cline{2-8}
&  Deeplab-VGG16            &  81.9 & 75.5&70.9 & 77.8 & \cellcolor{gray!25}75.6  & \cellcolor{gray!25}{\bf78.1} \\ \cline{2-8}
&  ResNet-101                   & 85.7 & 78.4&77.7 &82.0  & \cellcolor{gray!25}80.6 & \cellcolor{gray!25}{\bf82.2} \\ \hline\hline
\multirow{ 3}{*}{\begin{sideways}\parbox{10ex}{\centering trimap\\8 pixels}\end{sideways}} &  Deeplab-MSc-largeFOV & 75.0 & 69.5 & 60.3 & {\bf72.5} & \cellcolor{gray!25}68.4&\cellcolor{gray!25}71.4 \\ \cline{2-8}
&  Deeplab-VGG16            & 76.9 & 70.4&64.1 & 72.0 & \cellcolor{gray!25}69.0 & \cellcolor{gray!25}{\bf72.4} \\ \cline{2-8}
&  ResNet-101                   & 81.5 & 73.8&71.2 &76.7  & \cellcolor{gray!25}74.6 & \cellcolor{gray!25}{\bf77.0} \\ \cline{2-8}
\end{tabular}
\vspace{1ex}
\caption{Pixel-wise accuracy on \textit{val} set of PASCAL 2012. Top 3 rows: accuracy over all pixels. Middle 3 rows: accuracy for pixels within 16 pixels away from semantic boundaries. Bottom 3 rows: accuracy for pixels within 8 pixels aways from semantic boundaries. Pixels closer to boundaries are more likely to be mislabeled. Our ADM scheme improves over GD for grid CRF loss consistently for different networks. Note that weak supervision with our approach is almost as good as full supervision.}
\label{tab:accuracy}
\end{table*}

\subsection{Shortened Scribbles}
\label{sec:shortenedscribbles}
\begin{figure*}[h!]
    \centering
    \includegraphics[width=.45\textwidth]{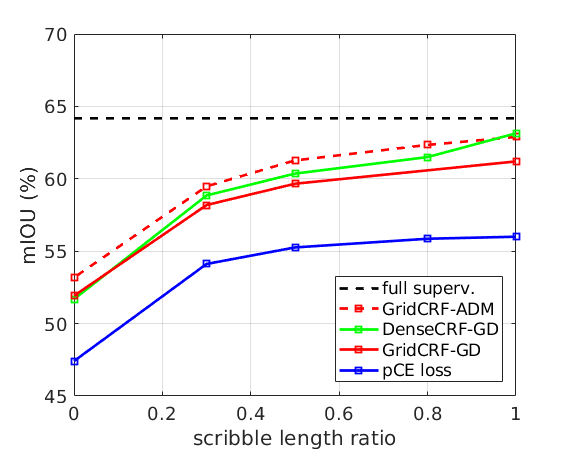}
    \includegraphics[width=.475\textwidth]{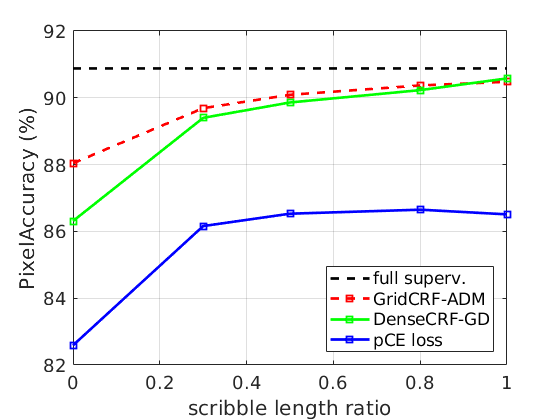}
   \caption{Experiment results of training with shorter scribbles with variants of regularized loss approaches. The results are for Deeplab-MSc-largeFOV. We report mIOU (left) and pixel-wise accuracy (right).}
   \label{fig:shortscribbles}
\end{figure*}

\vspace{-1ex}
Following the evaluation protocol in ScribbleSup \cite{scribblesup}, we also test our regularized loss approaches training with shortened scribbles. We shorten the scribbles from the two ends at certain ratios of length. In the extreme case, scribbles degenerate to clicks for semantic objects. We are interested in how the weakly-supervised segmentation methods degrade as we reduce the length of the scribbles. We report both mIOU and pixel-wise accuracy. As shown in Fig.~\ref{fig:shortscribbles}, our ADM for the grid CRF loss outperforms all competitors giving significantly better mIOU and accuracy than GD for the grid CRF loss. ADM degrades more gracefully than  the dense CRF as the supervision weakens.

The grid CRF has been overlooked in regularized CNN segmentation currently dominated by the dense CRF as either post-processing or trainable layers. We show that for weakly supervised CNN segmentation, the grid CRF as the regularized loss can give segmentation at least as good as that with the dense CRF. The key to minimizing the grid CRF loss is better optimization via ADM rather than gradient descent. Such competitive results for the grid CRF loss confirm that it has been underestimated as a loss regularizer for neural network training, as discussed in Sec.~\ref{sec:intro}.

It has not been obvious whether the grid CRF as a loss is beneficial for CNN segmentation. We show that straightforward gradient descent for the grid CRF does not work well. 
Our technical contribution on optimization helps to reveal the limitation and advantage of the grid CRF vs dense CRF models. The  weaker regularization properties, as discussed in Sec.~\ref{sec:shallow}, of the dense CRF and our experiments favors the grid CRF regularizer compared to the dense CRF.





%
%
%

\section{Conclusion}
Gradient descent (GD) is the default method for training neural networks. Often, loss functions and network architectures are designed to be 
amenable to GD. The top-performing weakly-supervised CNN segmentation \cite{NCloss:CVPR18,Rloss:ECCV18} 
is trained via regularized losses, as common in weakly-supervised deep learning \cite{weston2012deep,goodfellow2016deep}. 
In general, GD allows any differentiable regularizers. However, in shallow image segmentation it is known that generic GD 
is a substandard optimizer for (relaxations of) standard robust regularizers, \eg grid CRF.

Here we propose a general splitting technique, ADM, for optimizing regularized losses. It can take advantage of many existing efficient 
regularization solvers known in shallow segmentation. In particular, for grid CRF our ADM approach using $\alpha$-expansion solver achieves significantly better optimization quality compared to GD. With such ADM optimization, training with grid CRF loss achieves the-state-of-the-art in weakly supervised CNN segmentation. We systematically compare grid CRF and dense CRF losses from modeling and optimization perspectives. Using ADM optimization, the grid CRF loss achieves CNN training favourably comparable to the best results with the dense CRF loss. 
Our work suggests that in the context of network training more attention should be paid to optimization methods beyond GD.

In general, our ADM approach applies to many regularized losses, as long as there are efficient solvers for the corresponding regularizers. 
This work is focused on ADM in the context of common pairwise regularizers. Interesting future work is to investigate 
losses with non-Gaussian pairwise CRF potentials and higher-order segmentation regularizers, 
\eg $P^n$ Potts model \cite{kohli2009robust}, curvature \cite{Nieuwenhuis_2014_CVPR}, and kernel clustering \cite{Shi2000,Tang2019kernel}. 
Also with ADM framework, we can explore other optimization methods \cite{kappes2015comparative} besides $\alpha$-expansion for various kinds of regularized losses in segmentation. Our work bridges optimization method in "shallow" segmentation and loss minimization in deep CNN segmentation.



\small
\bibliographystyle{ieee}   
\bibliography{cvpr19_PottsADM}   

\end{document}